\documentclass[sigconf]{acmart}

\AtBeginDocument{%
  }

\setcopyright{none}
\copyrightyear{2026}
\acmYear{2026}
\acmDOI{}
\acmConference[HEAL@CHI '26]{3rd HEAL Workshop at CHI Conference on Human Factors in Computing Systems}{April 13--17, 2026}{Barcelona, Spain}

\acmBooktitle{3rd HEAL Workshop at CHI Conference on Human Factors in Computing Systems, Barcelona, Spain}
\acmISBN{}

\usepackage{tabularx}
\usepackage{booktabs}
\usepackage{bbm}
 \usepackage{float}
\usepackage{multirow}
\usepackage{graphicx}
\usepackage{tablefootnote}
\usepackage{amsmath}

\usepackage{amssymb}
\usepackage[ruled, vlined]{algorithm2e}
\usepackage{xcolor}
\usepackage{makecell} 
\usepackage[normalem]{ulem}
\usepackage{array}
\usepackage{arydshln}
\usepackage{algpseudocode}
\usepackage{subfigure}
\usepackage{diagbox}
\usepackage{makecell}
\usepackage[most]{tcolorbox}
\usepackage{listings}
\usepackage{adjustbox}
\usepackage{tikz}
\usepackage{xcolor}

\providecommand{\doi}[1]{\url{https://doi.org/#1}}
\begin{document}

\title{Casual as an Anchor: Resolving Supervision Misalignment in Formality Transfer Dataset}

\author{Hyojeong Yu}
\authornote{Both authors contributed equally to this research.}
\email{hyoj.yu@snu.ac.kr}
\affiliation{%
  \institution{Seoul National University}
  \city{Seoul}
  \country{Korea}
}
\author{Hyukhun Koh}
\authornotemark[1]
\email{hyukhunkoh-ai@snu.ac.kr}
\affiliation{%
  \institution{Seoul National University}
  \city{Seoul}
  \country{Korea}
}

\author{Minsung Kim}
\email{kms0805@snu.ac.kr}
\affiliation{%
  \institution{Seoul National University}
  \city{Seoul}
  \country{Korea}
}

\author{Kyomin Jung}
\authornote{Corresponding Author}
\email{kjung@snu.ac.kr}
\affiliation{%
  \institution{Seoul National University}
  \city{Seoul}
  \country{Korea}
}


\begin{abstract}
    Formality transfer is commonly framed as a symmetric bidirectional task between informal and formal registers. We argue that this framing conceals a supervision design flaw in existing benchmarks such as GYAFC: binary human rewrites encode relative stylistic shifts rather than absolute human notions of formality. Consequently, models learn to generate pseudo-formal outputs that satisfy benchmark labels while failing to produce genuinely formal language. We quantify this misalignment by re-evaluating benchmark "formal" labels under a human-aligned definition of formality, revealing substantial discrepancies that propagate to consistent informal$\rightarrow$formal failures across model families. To address this issue, we reconceptualize formality transfer as a graded dimension rather than a binary attribute. To operationalize this view, we introduce a three-level spectrum---informal, casual, and formal---where casual serves as an explicit intermediate state that clarifies supervision signals. Based on this framework, we introduce 3LF, a dataset providing parallel supervision across all three levels. Training on 3LF substantially reduces informal$\rightarrow$formal failures and improves alignment with human perception. For example, GPT-4.1-nano improves from 0.06 to 0.88 F1 in the informal$\rightarrow$formal direction despite 3LF being significantly smaller than GYAFC. We further demonstrate that these gains cannot be reproduced through in-context learning alone and provide qualitative analyses of ambiguity-driven errors and meaning distortions. Overall, our findings demonstrate how supervision design shapes stylistic alignment and highlight the importance of alignment-aware benchmark construction in controllable text generation.
\end{abstract}




\maketitle

\small
\textbf{Dataset Availability:}
\url{https://huggingface.co/datasets/stellahj/3lf}
\normalsize

\section{Introduction}
Text style transfer, particularly formality transfer, is a central task in controllable text generation. It involves rewriting a sentence into a target stylistic register---such as converting informal text into a formal tone---while preserving its original meaning~\cite{pavlick-tetreault-2016-empirical,rao-tetreault-2018-dear,briakou-etal-2021-ola,mukherjee2024textstyletransferintroductory}. Traditionally, formality transfer is treated as a binary transformation between informal and formal registers. However, our comprehensive evaluation reveals a persistent directional asymmetry: state-of-the-art models consistently underperform when converting informal text to formal language, while performing relatively well in the reverse direction~\cite{liu-etal-2024-step, Lai2024StyleSpecificNF,saakyan-muresan-2024-iclef,toshevska-etal-2025-style}.

We show that this asymmetry arises from a misalignment between benchmark supervision and human perceptions of formality. Existing datasets such as GYAFC~\cite{rao-tetreault-2018-dear} rely on binary human rewrites that primarily emphasize surface-level corrections rather than producing genuinely formal expressions characterized by hedging, nominalization, or passive constructions~\cite{Heylighen_and_Dewaele,Biber_1988,Biber_Conrad_2009}. As a result, this supervision misalignment collapses distinct stylistic intents into a single "formal" label, encouraging models to learn relative shifts instead of aligning to human-defined formality. This simplification obscures the fact that formality naturally forms a graded spectrum-informal, casual, and formal-where casual serves as an intermediate register that is grammatically clean but less rigid than formal text.

To address this misalignment, we introduce 3LF, a dataset explicitly designed to model formality as a three-level spectrum (informal--casual--formal), providing aligned sentence triples across all levels. By making the intermediate casual state explicit, 3LF offers a more interpretable and alignment-aware supervision signal grounded in clear linguistic criteria.

We evaluate multiple model families (GPT-4.1-nano, Flan-T5-Large, and DeepSeek-Distill-Qwen-1.5B) under controlled training settings, comparing 3LF with traditional binary supervision. Across all models, training on 3LF substantially improves informal$\rightarrow$formal performance-a direction where binary-trained models consistently fail. These gains cannot be replicated through in-context learning alone, and are supported by qualitative analyses of meaning distortions and ambiguity-driven errors. These findings suggest that dataset supervision design plays a central role in shaping human-perceived stylistic quality, and therefore deserves greater research attention alongside advances in prompting and model scaling.

In summary, our contributions are threefold:
\begin{itemize}
    \item We systematically analyze directional asymmetry in formality transfer and show that persistent informal$\rightarrow$formal failures stem from structural misalignment in benchmark supervision.
    
    \item We introduce a theoretically grounded three-level formality spectrum (informal--casual--formal) and present 3LF, a carefully constructed dataset that provides explicit and interpretable supervision for each formality level.

    \item We demonstrate that alignment-aware training with 3LF significantly improves informal$\rightarrow$formal transfer and yields outputs that better reflect human-defined formality.
\end{itemize}{}

\section{Related Work}
Formality style transfer traditionally involves converting text between formal and informal styles, often using datasets like GYAFC~\cite{rao-tetreault-2018-dear}. Recent critiques highlight that these benchmarks focus on superficial modifications, inadequately representing true linguistic formality. \citet{liu-etal-2024-step,lai-etal-2024-style} emphasize these shortcomings, noting that even advanced models frequently generate outputs with informal elements. \citet{toshevska-etal-2025-style} suggest enhancing models through knowledge graphs to achieve deeper linguistic transformations, though rigorous human evaluation remains needed. Further research~\cite{mukherjee2024textstyletransferintroductory,saakyan-muresan-2024-iclef,mukherjee-etal-2025-evaluating} argues for more robust evaluations that include intermediate stylistic states, criticizing overly simplistic binary formality definitions. Recent proposals advocate for expert-guided intermediate feedback to improve transformations, especially from informal to formal styles, yet empirical validation is still necessary.

Overall, these studies emphasize the need for richer datasets and nuanced evaluation approaches to address limitations in existing benchmarks like GYAFC.

\section{Revisiting Formality in Existing Benchmarks}
\label{sec:reevaluation}
To analyze the asymmetry in formality transfer, we revisit existing datasets and uncover annotation inconsistencies that introduce misaligned supervision signals and compromise their reliability as evaluation standards. While GYAFC~\cite{rao-tetreault-2018-dear} has become the standard benchmark for formality transfer evaluation, our systematic analysis reveals fundamental issues with its formality annotations that may explain the observed directional bias in model performance.

\subsection{Linguistic Definition of Formality Spectrum}
\label{sec:formal_def}

First, as pointed out by~\citet{yang2025steeringlargelanguagemodels}, there is a concern about whether GYAFC truly contains formal states. Therefore, to rigorously define the different levels of formality, we refer to the theoretical, decontextualized definition of formal expressions~\cite{Heylighen_and_Dewaele}. Informal and formal expressions are determined by the frequency of non-deictic words present in the given sentence. Non-deictic words include nouns, adjectives, prepositions, and articles. More use of non-deictic words results in a more passive, formal tone. In contrast, a higher proportion of deictic words such as pronouns, verbs, adjectives, and interjections, results in a more direct, informal tone. However, it is mentioned that formality lies on a continuum, and that all linguistic expressions will be situated somewhere in between the two extremes. We define this ambiguous in-between style as ``casual'' tone.

To rigorously define the levels of formality, we draw on the theoretical and decontextualized definition of formal expressions proposed by~\citet{Heylighen_and_Dewaele}. Based on such definitions, we characterize the sequences along a formality spectrum---Informal, Casual, and Formal---by identifying representative linguistic features such as:

\begin{itemize}
    \item \textbf{Informal:} Characterized by the presence of slang, netspeak, interjections, emojis, non-standard spelling, and grammatical errors. This tone resembles spontaneous conversation in online settings.
    \begin{quote}
        \textit{``LOL that was sooo weird. idk what just happened but omg O\_O''}

    \end{quote}

    \item \textbf{Casual:} Uses contractions, abbreviations, and direct address (e.g., "you", "hey") but avoids overtly informal elements such as emojis or slang. It is relaxed yet grammatically clean.
    \begin{quote}
        \textit{"Hey, I'm not sure what happened, but it was quite weird."}
    \end{quote}

    \item \textbf{Formal:} Employs hedging phrases (e.g., "it appears that", "may suggest"), nominalization, and passive constructions. This tone emphasizes objectivity and detachment.
    \begin{quote}
        \textit{"It appears that an unexpected event occurred, the nature of which remains unclear."}
    \end{quote}
\end{itemize}

\begin{figure}[h]
    \centering
    \includegraphics[height=0.97\linewidth]{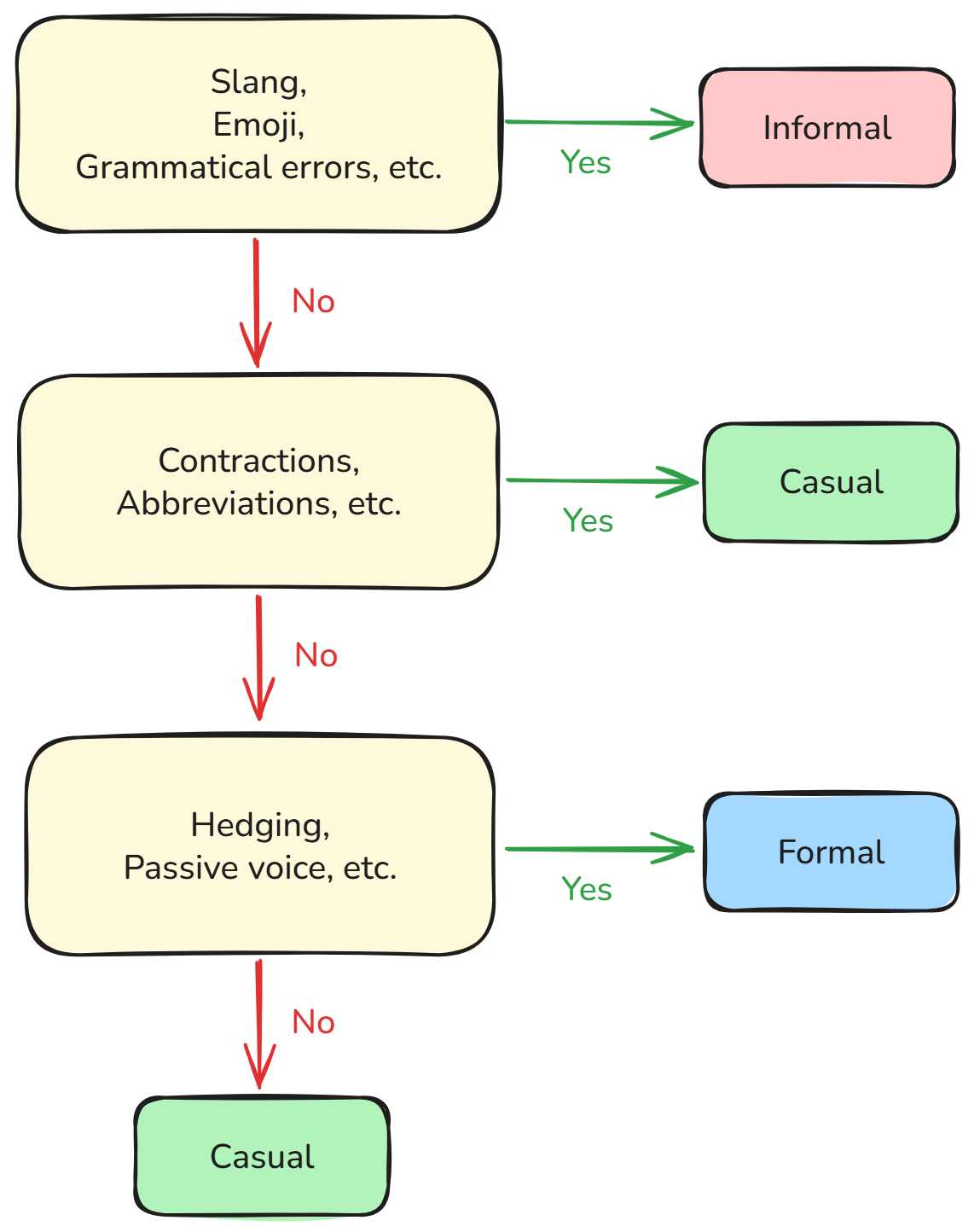}
    \caption{Rule-based decision tree for classifying sentence style.}
    \label{fig:tree}
    \vspace{-0.6cm}
\end{figure}

To translate these linguistic principles into a consistent annotation framework, we design a rule-based decision tree, illustrated in Figure~\ref{fig:tree}.

\subsection{Binary Formality Labels as Relative Preferences: Evidence from GYAFC}

Building on our definition of formality as a semantic continuum, we systematically examine the limitations of traditional binary-oriented benchmarks and classifiers. Specifically, we conduct a three-stage evaluation to assess the quality of GYAFC's "formal" labels. First, we apply a traditional formality classifier~\cite{dementieva-etal-2023-detecting} (See details of classifier construction in Appendix~\ref{appx:cls_choice}) to all sequences labeled as "formal" in GYAFC (approximately 100,000 sequences), achieving a classification accuracy of 91.1\%. However, this high score stems from classifiers trained on traditional datasets. To perform a more stringent evaluation, we employ theoretically grounded criteria to verify whether the sequences are truly formal. Due to resource limitations, we sample 1,000 examples and re-evaluate them using an LLM-based classifier, specifically GPT-4o (See Appendix Table~\ref{tab:formal-informal} for details), according to the definition in Section~\ref{sec:formal_def}. The results reveal a striking discrepancy: only 98 sequences (9.8\%) meet strict formality criteria, while 902 sequences (90.2\%) are reclassified as informal or casual. This massive discrepancy---from 91.1\% to 9.8\% formal classification---reveals a fundamental flaw in the human annotation design of the benchmark, where labels fail to align with theoretically grounded notions of formality.

A detailed investigation by three human annotators further confirms that the dataset frequently conflates casual or semi-formal text with genuinely formal language, often depending on the domain. The annotation process was reliable, with consistently high inter-annotator agreement (Fleiss' $\kappa > 0.7$ across all tasks), and disagreements were resolved by majority vote. For instance, the sentence \textit{"It is very similar to the age-old question, 'What if my blue is your red?'"} contains the term age-old, which might be considered formal in a general sense, but in professional domains such as OpenReview, it may not qualify as formal. Similarly, \textit{"Blue is my favorite color."} expresses a personal opinion and would be considered informal in the context of an official statement.

\subsection{Binary Annotation Induces Minimal-Edit Bias}
This discrepancy can be traced to GYAFC's annotation setup, where annotators rewrite informal sentences into formal versions. However, human evaluation of formal references in GYAFC reports an average score of 0.38 on a -3 to 3 scale~\cite{rao-tetreault-2018-dear}, indicating that these targets occupy a limited region of the stylistic spectrum. Rather than representing genuinely formal language, the rewrites often reflect relative shifts away from informality, operationalizing formality through the removal of surface markers rather than the incorporation of register-specific features. In broader text style transfer research, the lack of standardized evaluation procedures and inconsistent human evaluation practices have been shown to impede reliable style judgments, with automated metrics often poorly aligned with human perception of stylistic quality~\cite{ostheimer2023standardizationvalidationtextstyle}.

This relativistic framing---where formality is defined in relation to the input rather than through a theoretically grounded criterion~\cite{yang2025steeringlargelanguagemodels}---undermines the reliability of downstream evaluation and encourages supervision misalignment. Models trained under such benchmarks tend to produce pseudo-formal outputs that satisfy skewed criteria without achieving true formality. These observations motivate the need for an explicit stylistic anchor that defines formality beyond relative rewrites and provides stable reference points across intermediate states---an insight that directly informs the design of our three-level dataset.

\section{3LF Dataset Construction}

To resolve the conflation of casual or semi-formal text with genuinely formal language, we require a new dataset to deal with the directional asymmetry in performance across formality transfer. In particular, we treat the casual register as a human-interpretable anchor between informal and formal styles, enabling the construction of unambiguous alignment targets. To this end, we rewrite casual sentences into both formal and informal variants, forming rigorously aligned style triples. The following section details our methodology for creating the 3-Level Formality(\textbf{3LF}) dataset.

\subsection{Design Principle: Casual as a Stylistic Anchor}

\begin{figure*}[t]
    \centering
    \includegraphics[width=0.98\textwidth,height=0.55\textheight,keepaspectratio]{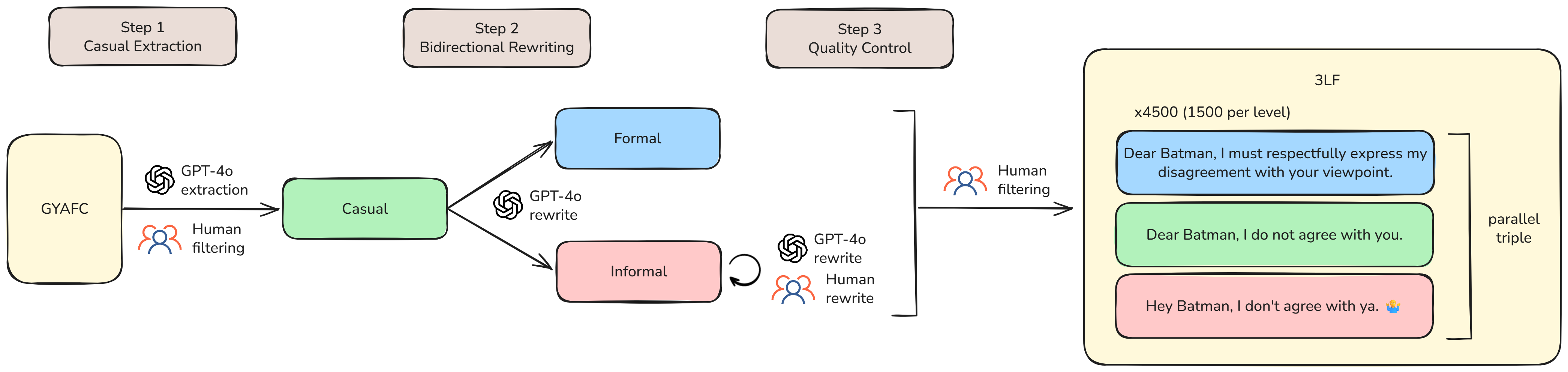}
    \caption{Construction pipeline of 3LF. Casual sentences are first identified from GYAFC using LLM-assisted filtering under fixed linguistic criteria. Each sentence is then rewritten into formal and informal variants within a human-in-the-loop pipeline, where human revision is applied at every stage to ensure alignment consistency and annotation quality.
}
    \label{fig:3lf_pipeline}
\end{figure*}

When constructing a dataset with a larger stylistic gap, direct transformation between informal and formal registers poses a significant challenge, as models often struggle to perform such large stylistic shifts while preserving the original meaning. Prior work has similarly noted that effectively adjusting stylistic attributes without compromising core content remains a central difficulty in the field~\cite{kong-etal-2025-neuron, mukherjee2024textstyletransferintroductory}. Also, informal sentences frequently exhibit syntactic incompleteness, omitting essential components such as subjects or predicates (e.g., "unless it's with the wrong man."), resulting in underspecified semantic content. Formalizing such expressions requires reconstructing implicit structure and supplying missing contextual and structural information, rather than merely adjusting surface style. As a result, informal$\rightarrow$formal transfer is not simply the reverse of formal$\rightarrow$informal rewriting; the two directions differ in their structural and informational demands. Treating them as symmetric tasks therefore obscures a fundamental asymmetry embedded in the transformation itself. In contrast, the casual register occupies an intermediate position along a graded dimension of formality. We hypothesize that such a casual anchor reduces the requirement for semantic inference over underspecified informal inputs, thereby disentangling content completion from stylistic transformation.

Starting from the casual anchor, a formal variant can be generated by selectively amplifying formal linguistic features---such as hedging, nominalization, and passivization---while suppressing informal elements. Conversely, an informal variant can be obtained by removing formal features and intensifying informal characteristics. This anchored formulation reduces the risk of meaning distortion and enables more reliable construction of stylistically distinct yet semantically aligned sentence variants, effectively decomposing a large and ambiguous shift into two smaller, more tractable transitions.

\subsection{Construction Pipeline}

We utilize GPT-4o for dataset construction within a human-in-the-loop pipeline designed to ensure annotation quality. Human revision is applied at every stage of LLM-assisted rewriting, and all annotators involved possess advanced proficiency in English. (See Appendix~\ref{appx:cls_choice}) Annotators were provided with our formality decision tree illustrated in Figure~\ref{fig:tree} to guide all revision processes. The entire construction pipeline is shown in Figure~\ref{fig:3lf_pipeline}.

\textbf{Training Set} We use the training split of GYAFC, a standard benchmark for formality style transfer. Based on our formality decision tree, we use GPT-4o as an automatic judge to extract sentences exhibiting casual tone from the original corpus. Our prompt follows the evaluation template introduced in~\citet{koh-etal-2024-llms}. (See Appendix~\ref{appx:3fl}) Then, we instruct the LLM to rewrite each extracted casual sentence into both formal and informal variants, thereby constructing parallel and explicitly aligned data. Human annotators verify whether each sentence conforms to the categories defined in Figure~\ref{fig:tree}. For the informal rewrites, we apply multiple rounds of targeted revision using curated prompts, along with human verification to ensure sufficiently strong informal signals. In total, we craft 4500 sequences, 1500 samples for formal, casual, informal respectively.

To verify the effectiveness of the 3LF dataset construction process, we also create a NAIVE-3LF set for comparison. We sample 4,500 informal sentences from the GYAFC corpus and rewrite them directly into formal style using GPT-4o, without employing the intermediate casual state construction approach.

\textbf{Test set} For evaluation, we construct a test set spanning two formality levels: informal and formal. The test data are drawn from two sources: GYAFC, derived from Yahoo Answers, and the Pavlick dataset \cite{pavlick-tetreault-2016-empirical}, which covers four domains-news, email, answers, and blogs-and is skewed toward more formal language. We sample 200 informal instances from the GYAFC test split and 200 formal instances from the Pavlick test split. Because the Pavlick dataset provides formality scores ranging from -3 to +3, we retain only examples with positive average scores to ensure a high degree of formality.
Following human evaluation of the sampled examples, the final test set comprises 400 sentences, evenly distributed with 200 examples. All annotations were conducted following the procedure illustrated in Figure~\ref{fig:tree}.

\subsection{Dataset Quality and Integrity}

We further assess potential data leakage by measuring lexical overlap between each training dataset (GYAFC, NAIVE, and 3LF) and the test set. Specifically, we report n-gram overlap statistics (from 1-gram to 5-gram) to quantify surface-level similarity between training and test instances.

\begin{table}[h]
\small
\resizebox{\linewidth}{!}{
\begin{tabular}{lccccc}
\toprule
\textbf{Dataset} & \textbf{1-gram} & \textbf{2-gram} & \textbf{3-gram} & \textbf{4-gram} & \textbf{5-gram} \\
\midrule
\textbf{GYAFC} & 0.771 & 0.478 & 0.207 & 0.061 & 0.016 \\
\textbf{NAIVE} & 0.380 & 0.164 & 0.024 & 0.003 & 0.000 \\
\textbf{3LF} & 0.401 & 0.193 & 0.036 & 0.005 & 0.001 \\
\bottomrule
\end{tabular}
}
\caption{N-gram Overlap Statistics}
\label{tab:ngram}
\vspace{-0.9cm}
\end{table}

\begin{table}[h]
\centering
\small
\begin{tabular}{ll|cc}
\toprule
\multicolumn{2}{l|}{\textbf{Dataset}}& \textbf{Characters} & \textbf{Words} \\
\midrule
\multirow{2}{*}{\textbf{GYAFC}} & Formal & 51.34 & 10.30 \\
& Informal & 55.87 & 10.97 \\
\midrule
\multirow{3}{*}{\textbf{3LF}} & Formal & 80.07 & 13.79 \\
& Casual & 53.19 & 10.43 \\
& Informal & 49.19 & 9.94 \\
\bottomrule
\end{tabular}
\caption{Sentence-Level Statistics on GYAFC and 3LF}
\vspace{-0.7cm}
\label{tab:sentence-level}
\end{table}

As shown in Table~\ref{tab:ngram}, lexical overlap between the training and test sets is limited at higher n-gram orders: while GYAFC exhibits a relatively high 1-gram overlap (0.771), overlap drops sharply for longer sequences (5-gram = 0.016). Both NAIVE and 3LF demonstrate substantially lower overlap across all n-gram orders, with near-zero overlap beyond trigrams.

These results indicate that the test set is not trivially recoverable from the training data and that performance gains observed in our experiments cannot be attributed to surface-level lexical memorization. Overall, 3LF maintains low lexical redundancy with the evaluation set while preserving sufficient linguistic diversity for robust training.

Additionally, sentence-level statistics (Table~\ref{tab:sentence-level}) show that 3LF exhibits clear stratification across levels, with formal sentences substantially longer and lexically richer, consistent with established linguistic characterizations of formal registers~\citep{Heylighen_and_Dewaele, Biber_Conrad_2009}.

\section{Experiments}

To assess the effectiveness of our 3LF dataset as supervision for formality-aware generation, we conduct controlled experiments on a generative formality transfer task. Specifically, we conduct controlled experiments across diverse model families to examine whether training with 3LF leads to more reliable and well-grounded formality transformations. This section details the experimental setup and evaluation protocol.

\subsection{Experimental Setup}
For training, we consider three settings: (i) a dataset incorporating the introduced casual state (\textbf{3LF}), (ii) a dataset without it (\textbf{NAIVE}), and (iii) the baseline dataset (\textbf{GYAFC}).

To investigate the impact of dataset quality on generation performance, we fine-tune three models: GPT-4.1-nano, Flan-T5-Large, and DeepSeek-Distill-Qwen-1.5B---across all three datasets and compare their results. We used customized prompts for each model, as shown in Appendix ~\ref{appx:generation}.

Next, we evaluate generation quality using a combination of automatic and human-centered metrics. For formality assessment, we compute precision, recall, and F1 scores for each rewriting direction (informal-to-formal and formal-to-informal), along with overall accuracy across both directions.

To further assess the quality of generated outputs, we measure fluency and meaning preservation, focusing only on sentences that exhibit the correct target formality.
Fluency is evaluated automatically using GPT-4o, which assigns an integer score between 0 and 5 based on sentence naturalness and grammar. Our evaluation prompt is based on recent LLM-based style transfer evaluation templates~\cite{ostheimer2023standardizationvalidationtextstyle,mukherjee-etal-2025-evaluating,pauli-etal-2025-mind} with temperature 0. Following \citet{koh-etal-2024-llms}, we further incorporate an explicit definition-based rubric for formality, ensuring that the evaluator applies theoretically grounded criteria rather than relying on implicit stylistic preferences (See Appendix~\ref{appx:generation}). For meaning preservation, three annotators independently determine whether the original and rewritten sentences convey the same meaning. Final labels are decided by majority vote. Due to the low fluency of outputs generated by DeepSeek-1.5B, we restrict annotation to generations from GPT-4.1-nano and T5-large, which consistently demonstrate higher fluency.

\begin{table*}[t]
\centering
\small
{
\renewcommand{\arraystretch}{1.2}
\begin{tabular}{
    ll
    !{\vrule width 0.8pt}
    ccc
    !{\vrule width 0.4pt}
    ccc
    !{\vrule width 0.4pt}
    ccc
}
\toprule
\multirow{2}{*}{\textbf{Model}} &
\multirow{2}{*}{\textbf{Dataset}} &
\multicolumn{3}{c!{\vrule width 0.4pt}}{\textbf{F$\rightarrow$I}} &
\multicolumn{3}{c!{\vrule width 0.4pt}}{\textbf{I$\rightarrow$F}} &
\multirow{2}{*}{\textbf{Acc.}} &
\multirow{2}{*}{\textbf{Fluency}} &
\multirow{2}{*}{\shortstack{\textbf{Meaning} \\ \textbf{Preservation}}}
\\
\cmidrule(lr){3-5}\cmidrule(lr){6-8}
& & P & R & F1 & P & R & F1 & ~\\
\midrule
\multirow{3}{*}{GPT-4.1-nano} 
  & GYAFC  & 0.49 & 0.93 & 0.64 & 0.33 & 0.04 & 0.06 & 0.4825 & 3.5759 & 0.8100\\
  & NAIVE  & 0.81 & 0.88 & 0.84 & 0.86 & 0.80 & 0.83 & 0.8350 & 3.9162 & 0.8225\\
  & 3LF   & 0.83 & 0.98 & 0.90 & 0.98 & 0.81 & 0.88 & \textbf{0.8950} & \textbf{4.2212} & \textbf{0.8525}\\
\hdashline
\multirow{3}{*}{T5-large} 
  & GYAFC  & 0.37 & 0.57 & 0.45 & 0.02 & 0.01 & 0.01 & 0.2925 & 4.3333 & \textbf{0.8750}\\
  & NAIVE  & 0.40 & 0.56 & 0.47 & 0.27 & 0.16 & 0.20 & 0.3600 & 4.5069 & 0.6775\\
  & 3LF   & 0.46 & 0.36 & 0.41 & 0.47 & 0.56 & 0.51 & \textbf{0.4650} & \textbf{4.5806} & 0.6900\\
\hdashline
\multirow{3}{*}{DeepSeek-1.5B} 
  & GYAFC  & 0.53 & 0.99 & 0.69 & 0.92 & 0.12 & 0.20 & 0.5525 & 1.5385 & -\\
  & NAIVE  & 0.64 & 0.76 & 0.70 & 0.71 & 0.57 & 0.63 & 0.6675 & 2.1386 & -\\
  & 3LF   & 0.72 & 1.00 & 0.84 & 1.00 & 0.61 & 0.76 & \textbf{0.8075} & \textbf{3.9164} & -\\
\bottomrule
\end{tabular}%
}
\caption{Bidirectional style transfer results. Accuracy is averaged over both directions. F$\rightarrow$I: formal$\rightarrow$informal; I$\rightarrow$F: informal$\rightarrow$formal.}
\label{tab:model-bidirectional}
\vspace{-0.7cm}
\end{table*}

\subsection{Formality Transfer Results}

As shown in Table~\ref{tab:model-bidirectional}, a consistent pattern emerges across all three models: fine-tuning on 3LF yields substantial improvements in formality accuracy, with the most pronounced gains in the informal-to-formal (I$\rightarrow$F) direction. For instance, GPT-4.1-nano achieves an I$\rightarrow$F F1 score of 0.88 when trained on 3LF, compared to 0.06 on GYAFC and 0.83 on NAIVE---an absolute improvement of +0.82 over the GYAFC baseline. This sharp contrast indicates that the primary weakness of existing benchmarks lies in the insufficient supervision for genuine formal generation. By anchoring stylistic transformations to a shared intermediate register, 3LF introduces a cognitively grounded reference point along the formality continuum, reducing polarity ambiguity and producing outputs that more closely align with human-perceived notions of formality.

Importantly, the improvements are not confined to a single direction. In the formal-to-informal (F$\rightarrow$I) setting, GPT-4.1-nano and DeepSeek-1.5B also show measurable F1 gains over the GYAFC baseline. This suggests that the bidirectional alignment induced by 3LF enhances stylistic controllability more broadly, stabilizing both mappings rather than disproportionately benefiting one.

Additionally, all models exhibit increased overall accuracy when trained on 3LF compared to NAIVE, validating the effectiveness of our data generation pipeline. Notably, these improvements also hold when compared against GYAFC, despite the smaller scale of the 3LF dataset. This result underscores that alignment quality and stylistic clarity can outweigh sheer data quantity in training effective style transfer models.

\subsection{Generation Quality}

We apply two metrics-fluency and meaning preservation-to evaluate generation quality. For fluency, GPT-4.1-nano and T5-large consistently achieve high scores above 3.5 regardless of the training dataset, indicating their robustness to data variation. In contrast, DeepSeek-1.5B exceeds this threshold only when trained on our 3LF dataset, while its fluency score drops below 2.0 when trained on GYAFC. These results suggest that training data quality has a substantial impact on the model's ability to produce fluent outputs.

We further evaluate meaning preservation for GPT-4.1-nano and T5-large. GPT-4.1-nano maintains consistently high scores across all datasets. The model trained on 3LF achieves the highest preservation score, indicating that the anchored alignment structure of our dataset helps the model preserve semantic content while performing non-trivial stylistic transformations.

T5-large achieves an even higher meaning preservation score of 0.8750 on the GYAFC dataset. However, a qualitative inspection of the generated outputs reveals that this high score is largely driven by the model's tendency to copy or minimally modify the input sentence rather than performing substantive rewriting. Consequently, the model exhibits almost no effective formality transfer, with overall accuracy dropping to as low as 0.29. In contrast, when trained on 3LF, the model achieves a substantially higher accuracy of 0.46, while maintaining competitive preservation scores, indicating more meaningful stylistic transformation rather than superficial retention.

Overall, these results indicate that effective formality transfer requires supervision that encourages non-trivial stylistic transformation while preserving semantic content. Datasets that provide explicit and well-aligned stylistic supervision---such as 3LF---enable models to achieve this balance, yielding fluent outputs with faithful meaning preservation and reliable formality control.

\subsection{Qualitative Analysis}
\paragraph{Coreference / Discourse-level distortion}
We further conduct qualitative analysis on generated outputs. For T5-large, we observe discourse-level shifts in deixis and coreference that subtly alter interpretation (e.g., "No way im 5'4 and he's 6'2" $\rightarrow$ "I am 5'4 and he is 6'2"). For GPT-4.1-nano, there are several types of meaning distortion. These include entity shifts (e.g., "good god how old are you" $\rightarrow$ "One may wonder about the age of the individual in question"), numerical inaccuracies, and subjective bias injection. These errors reflect failures to preserve speaker stance and referential structure, highlighting that stylistic rewriting interacts with pragmatic meaning rather than merely surface-level editing.

\paragraph{Underspecification in Informal Language}
A second class of errors stems from the inherent underspecification of informal language. Informal expressions frequently omit subjects, contextual grounding, or explicit propositional structure, making their semantic intent ambiguous. When rewriting such inputs into formal style, models must implicitly infer missing information, which introduces aleatoric uncertainty. As a result, the transformation task becomes entangled with content completion rather than purely stylistic modulation. This explains why informal-to-formal rewriting is particularly error-prone and motivates our use of a casual anchor to reduce semantic ambiguity prior to formal transformation. Representative examples for each error cases are provided in Appendix~\ref{appx:error}.

\subsection{Comparison with In-Context Learning}
\begin{table}[h]
\centering
\small
\begin{tabular}{lccc}
\toprule
\textbf{Model} &
\textbf{Accuracy} &
\makecell{\textbf{F1}\\ \textbf{(F$\rightarrow$I)}} &
\makecell{\textbf{F1} \\ \textbf{(I$\rightarrow$F)}} \\
\midrule
Zero-Shot & 0.52 & 0.16 & 0.66 \\
ICL-GYAFC & 0.54 & 0.67 & 0.23 \\
ICL-3LF & 0.76 & 0.80 & 0.69 \\
FT-GYAFC & 0.48 & 0.64 & 0.06 \\
FT-NAIVE & 0.83 & 0.84 & 0.83 \\
FT-3LF & 0.89 & 0.90 & 0.88 \\
\bottomrule
\end{tabular}
\caption{Comparison Across Supervision Strategies}
\vspace{-0.7cm}
\label{tab:icl}

\end{table}

To verify that the observed improvements stem from 3LF supervision rather than the inherent capability of the base model, we compare fine-tuning (FT) with in-context learning (ICL) on both 3LF and GYAFC. We additionally report zero-shot performance as a baseline. To assess the model's default stylistic prior, we remove any artificial role framing (e.g., "You are not an AI assistant...") and evaluate GPT-4.1-nano under a four-shot prompting setup. The four-shot demonstrations are sampled from each dataset according to an informal-targeted selection criterion.

As shown in Table~\ref{tab:icl}, ICL-3LF exhibits the same failure patterns observed in earlier experiments, including directional asymmetry and meaning distortions in informal-to-formal rewriting. While ICL partially mitigates these issues, fine-tuning on 3LF yields substantially more consistent improvements, reducing informal-to-formal failures and producing outputs that better align with the intended formality register. In contrast, fine-tuning on GYAFC performs even worse than ICL-GYAFC, which itself is comparable to zero-shot performance. We further observe that the zero-shot model struggles more with generating informal sentences than formal ones, a pattern we also encountered during dataset construction. Notably, performance on the I$\rightarrow$F direction declines with increased exposure to GYAFC supervision, suggesting that supervision derived from a narrowly defined binary benchmark may bias the model away from its original stylistic prior. These results further support our claim that relative binary supervision may provide insufficient signal for genuine formal generation.

\section{Conclusion}
This study challenges the long-standing binary perspective of formality transfer and demonstrates that effective style transformation requires modeling formality as a graded dimension. By introducing the 3LF dataset---explicitly incorporating an intermediate casual state---we present a structured framework that addresses the persistent informal$\rightarrow$formal collapse observed in prior benchmarks such as GYAFC. Beyond the specific case of formality transfer, our findings carry broader design implications. The way stylistic categories are operationalized during dataset construction fundamentally shapes model behavior and perceived competence. While much work has explored advances in model architecture, prompting strategies, and scale, comparatively less attention has been devoted to examining how supervision encodes linguistic constructs. Our results suggest that dataset design is not a peripheral engineering choice, but a central factor influencing alignment with human-perceived language categories. Accordingly, supervision pipelines should be treated as core research artifacts requiring theoretical grounding and empirical validation. We hope this work contributes to the development of more nuanced, faithful, and controllable text generation systems that better reflect linguistic reality and align more closely with human perception.

\begin{acks}
This work was supported by Institute of Information \& communications Technology Planning \& Evaluation(IITP) grant funded by the Korea government(MSIT) [No.RS-2023-
00229780, Development of Artificial Intelligence Technology for Process-focused Evaluation(Student's Learning Diagnosis)]. K. Jung is with ASRI, Seoul National University, Korea.
\end{acks}



\clearpage
\newpage
\appendix

\section{Evaluator Choices}
\label{appx:cls_choice}
\begin{table}[ht]
\centering
\footnotesize
\begin{tabular}{lccc}
\toprule
\textbf{Model} & \textbf{Formal} & \textbf{Informal} & \textbf{Total} \\
\midrule
Monolingual    & 0.952 & 0.903 & 0.912 \\
Multilingual   & 1.000 & 0.000 & 0.175 \\
LLM Evaluation  & 1.000 & 0.976 & 0.980 \\
\bottomrule
\end{tabular}
\caption{Performance comparison across formal and informal domains}
\label{tab:formal-informal}
\end{table}

Before exploring the inherent issues in traditional benchmark, we begin by evaluating classifiers on a widely used formality dataset, specifically selecting sequences with high annotator agreement (scores exceeding 2.5 or below -2.5)~\cite{pavlick-tetreault-2016-empirical}, to select an appropriate formality evaluator. We compare existing formality classifiers such as a monolingual classifier~\cite{10.1145/3678179} and a multilingual classifier~\cite{dementieva-etal-2023-detecting}. In addition, given recent advancements, LLM-based evaluations have emerged as promising alternatives for tasks requiring semantic understanding and application of nuanced definitions~\cite{koh-etal-2024-llms}. We therefore also experiment with an LLM-based evaluator (GPT-4o) with temperature 0, using carefully curated prompts grounded in the theoretical, decontextualized definition of formality proposed by \citet{Heylighen_and_Dewaele} (See Appendix~\ref{appx:llmeval}). As shown in Table~\ref{tab:formal-informal}, the LLM-based evaluation achieves the highest overall performance. Unless otherwise specified, subsequent experiments in this paper adopt the LLM-based formality classifier. In the following human evaluation, all annotations were performed by a group of professional annotators with advanced proficiency in English. The annotators have extensive experience in English-language NLP tasks, including contributions to multiple academic papers, and possess strong linguistic backgrounds. All annotators have demonstrated high English fluency through prior professional and academic work.

\section{Formality Transfer with Sentiment}
\label{appx:fts}

\begin{table}[ht]
\renewcommand{\arraystretch}{1.2}
\centering
\small
\resizebox{\linewidth}{!}{
\begin{tabular}{lcccc}
\toprule
\textbf{Domain} & \textbf{Cases} & \textbf{Positive Modulation} & \textbf{Negative Modulation} & \textbf{Total Rate} \\
\midrule
General       & 28 & 18 & 10 & 14.0\% \\
Review        & 35 & 18 & 17 & 17.5\% \\
Social/Chat   & 49 & 12 & 37 & 24.5\% \\
Professional  & 43 & 7  & 36 & 21.5\% \\
Real-time     & 40 & 24 & 16 & 20.0\% \\
\bottomrule
\end{tabular}
}
\caption{Sentiment Modulation across domain-specific generations}    
\vspace{-0.3cm}
\label{tab:sentiment-shift}
\end{table}

\subsection{Sentiment Softening Across Domains}
Throughout our experiments, we observe that formality transfer often results in a softening of the original sentences' emotional tone. 

This phenomenon warrants deeper investigation and discussion. Therefore, we analyze patterns of sentiment shifts in four different domains: review style, casual social media and chat-based, professional communication, and real-time and retrospective response. \textbf{Review-style} domains---including Amazon product reviews, Yelp restaurant critiques, and film evaluations---display informality through highly subjective, evaluative, and personalized language, typically accompanied by explicit sentiment polarity~\cite{hartmann-etal-2023-more}. Conversely, \textbf{casual social media and chat-based} platforms, such as Reddit discussions, Twitter interactions, and online news commentary, express informality via colloquialisms, contractions, emojis, and paralinguistic markers, reflecting spontaneous and immediate emotional responses~\cite{sun-etal-2024-toward}. In contrast, \textbf{professional communication} contexts---encompassing peer reviews on OpenReview, LinkedIn discourse, formal email correspondence, and enterprise Slack exchanges---commonly balance relational rapport with linguistic formality, characterized by lexical precision, syntactic clarity, and measured affective expression~\cite{Tyler01042005,petersonetal}. Finally, \textbf{real-time and retrospective response} domains, exemplified by live-stream commentary, event-specific discourse, and reflective social media posts, frequently employ intensified emotional expressions, compressed syntactic structures, and temporal anchoring strategies, typically realized through present-tense narration and emotive discourse markers~\cite{10.1145/2556195.2556261,deng-etal-2024-text}. 

\subsection{Empirical Analysis of Sentiment Shift}
For this analysis, we again use GPT-4o, assigning sentiment labels using the \url{siebert/sentiment-roberta-large-english} classifier~\cite{HARTMANN202375}. We also use domain-specific prompts as in Appendix~\ref{appx:prompt_failure}. As shown in Table~\ref{tab:sentiment-shift}, notable cases of sentiment softening occur consistently across all domains during formality transformation. The systematic sentiment modulation we observe in Table~\ref{tab:sentiment-shift} is not a side effect but a core mechanism explaining both the asymmetry problem and our framework's success.  

Formal language typically emphasizes politeness, social deference, and interpersonal distance over emotional expression~\cite{Brown_Levinson_1987, pavlick-tetreault-2016-empirical}, deliberately avoiding emotionally charged lexical items and overt subjective expressions, resulting in predominantly neutral or mildly positive sentiment patterns~\cite{aithal-tan-2021-positivity}. Conversely, informal language frequently employs colloquialisms, contractions, slang, and syntactic looseness---linguistic devices that inherently convey emotional stance and subjective opinions~\cite{Biber_Conrad_2009,Culpeper_2011}, making informal utterances naturally conducive to sentiment-laden content.

Due to their likelihood-based training objectives, LLMs inherently favor frequent co-occurrence observed in such real-world training data~\cite{brown2020languagemodelsfewshotlearners,10.1145/3442188.3445922}, and  \citet{shani2025tokensthoughtsllmshumans} further shows that LLMs emulate abstract cognitive patterns from human language use, leading to subtle but systematic shifts in affective expression during style transfer~\cite{han-etal-2022-balancing}.

While sentiment modulation is therefore a natural linguistic phenomenon, certain practical scenarios require careful consideration to prevent unintended communicative effects. In professional and formal contexts especially, it may be essential for users to preserve specific sentiment orientations despite stylistic transformations, particularly when communicative intent critically depends on maintaining the original emotional stance.

\section{Detailed Analysis on 3LF}

\subsection{Sentence-Level Statistics}
We report sentence length and word distribution statistics across formality levels in Table \ref{tab:sentence-level}. In GYAFC, formal sentences average 51.34 characters, compared to 55.87 for informal ones. In 3LF, the distributions are clearly stratified: informal (49.19 chars / 9.94 words), casual (53.19 / 10.43), and formal (80.07 / 13.79). These results confirm that formal outputs in 3LF are substantially longer and lexically richer than their informal and casual counterparts, aligning with established linguistic observations that formal registers favor nominalization, hedging, and syntactic elaboration. \citep{Heylighen_and_Dewaele, Biber_Conrad_2009}

\subsection{Fluency Evaluation}
For fluency metrics, we conducted additional experiments using the CoLA classifier (iproskurina/tda-roberta-large-en-cola). We measured the number of sentences classified as acceptable in each dataset: GYAFC yielded 342 out of 400, NAIVE 312 out of 400, and 3LF 341 out of 400. These results suggest that, although most outputs are broadly acceptable, CoLA is not sufficiently sensitive to capture fine-grained distinctions in fluency across datasets. This contrasts with our GPT-4o--based fluency evaluations, which provide clearer separation and align more closely with human judgments.

\onecolumn
\section{Prompt Examples}
\subsection{Evaluation Prompt}
\label{appx:llmeval}
\begin{tcolorbox}[title=Evaluation Prompt, colback=gray!15, colframe=black!60]
\textbf{<SYS>}

 You are the formality style transfer agent. Here are the definitions of Formal, Casual, Informal sequence respectively.
 
- Formal sequence : employs hedging phrases (e.g., "it appears that", "may suggest"), nominalization, and passive constructions. This tone emphasizes objectivity and detachment. 

- Casual sequence : uses contractions, abbreviations, and direct address (e.g., "you", "hey") but avoids overtly informal elements such as emojis or slang. It is relaxed yet grammatically clean.

- Informal sequence : is characterized by the presence of slang, netspeak, interjections, emojis, non-standard spelling, and grammatical errors. This tone resembles spontaneous conversation in online settings.

Evaluate the Target sequence, and nothing else.

\textbf{</SYS>}

\textbf{<INST>}

Evaluate the Target sequence.
If the sequence is formal, answer 1, and nothing else.
else the sequence is informal, answer 0, and nothing else.

Target sequence : {}

The answer is 

\textbf{</INST>}
\end{tcolorbox}

\subsection{3LF Prompt}
\label{appx:3fl}
\begin{tcolorbox}[title=Formality Classification Prompt, colback=gray!15, colframe=black!60]
You are not an AI assistant. You are a specialized formality classification machine that can only output three integer labels: 0, 1, or 2.
Label the following sentence on formality based on the presence of specific linguistic features. Strictly follow the following labeling rules:

Label 0:
- Assign Label 0 if the sentence contains **any** of the following: slang, netspeak, interjections, emojis, non-standard spellings, or grammatical errors.

Label 1:
- If none of the above features are present, assign Label 1 if the sentence includes **any** of: contractions, abbreviations, or direct address.

Label 2:
- If none of the above apply, and the sentence contains **any** of the following: hedging phrases, nominalizations, or passive voice --- assign Label 2.

Note:
- Label 1 should also be assigned to all the other sentences that do not contain strong stylistic features listed above.

Take a deep breath and think step by step.
\end{tcolorbox}

\begin{tcolorbox}[title=3LF Generation Prompt, colback=gray!15, colframe=black!60]
You are not an AI assistant. You are a specialized formality transfer machine that can only output rewritten sentences.
Rewrite the following casual sentence to an informal sentence based on the presence of specific linguistic features:

Informal:
- A sentence is informal if it contains **any** of the following: slang, netspeak, interjections, emojis, non-standard spellings, or grammatical errors.

Casual:
- A sentence is casual if none of the above features are present, and includes **any** of: contractions, abbreviations, or direct address.

Formal:
- A sentence is formal if none of the above apply, and it contains **any** of the following: hedging phrases, nominalizations, or passive voice.

Note:
- All the other sentences that do not contain strong stylistic features listed above are also considered casual.

Only answer with the rewritten sentence.
Take a deep breath and think step by step.
\end{tcolorbox}

\subsection{Generation Prompt}
\label{appx:generation}

\begin{tcolorbox}[title=GPT-4.1-nano, colback=gray!15, colframe=black!60]
You are not an AI assistant. You are a specialized sentence rewriting model.
Rewrite the given sentence to formal, based on the presence of specific linguistic features:

Informal: sentences containing slangs, netspeak, interjection, emoji, non-standard spelling, grammatical errors

Formal: sentences containing hedging phrase, nominalization, passive voice.

Only answer with the rewritten sentence.
Take a deep breath and think step by step.
\end{tcolorbox}

\begin{tcolorbox}[title=T5-large, colback=gray!15, colframe=black!60]
Definition:

  - Informal: A sentence is informal if it contains slangs, netspeak, interjections, emojis, non-standard spelling, or grammatical errors.
  
  - Formal: A sentence is formal if it contains hedging phrases, nominalizations, or passive voice.

  Instruction:
  
  Rewrite the following sentence to formal.

  Sentence: \{sentence\}
\end{tcolorbox}

\begin{tcolorbox}[title=Deepseek 1.5B, colback=gray!15, colframe=black!60]
You are not an AI assistant. You are a specialized sentence rewriting model.
Rewrite the given sentence to formal, based on the presence of specific linguistic features:

Informal: sentences containing slangs, netspeak, interjection, emoji, non-standard spelling, grammatical errors

Formal: sentences containing hedging phrase, nominalization, passive voice.

Only answer with the rewritten sentence.

Sentence: 
\end{tcolorbox}

\begin{tcolorbox}[title=Fluency Evaluation Prompt, colback=gray!15, colframe=black!60]
You are a specialized fluency scoring machine that can only output integer scores.

Score the following context on fluency on an scale from 0 to 5, where 5 is of the highest fluency.

Only focus on whether the sentence sounds natural and coherent.

Do not penalize an informal sentence for informal style features such as slang or contractions,
and do not penalize a formal sentence for formal style features such as hedging phrases or passive voice.

Take a deep breath and think step by step.

**Answer only with the integer value.**
\end{tcolorbox}

\subsection{Analysis Prompt}
\label{appx:prompt_failure}
\begin{tcolorbox}[title=System Prompt, colback=gray!15, colframe=black!60]
\textbf{<SYS>}
 You are the formality style transfer agent. Here are the definitions of Formal, Casual, Informal sentence respectively.
 
- Formal sentence : employs hedging phrases (e.g., "it appears that", "may suggest"), nominalization, and passive constructions. This tone emphasizes objectivity and detachment. 

- Casual sentence : uses contractions, abbreviations, and direct address (e.g., "you", "hey") but avoids overtly informal elements such as emojis or slang. It is relaxed yet grammatically clean.

- Informal sentence : is characterized by the presence of slang, netspeak, interjections, emojis, non-standard spelling, and grammatical errors. This tone resembles spontaneous conversation in online settings.

Make the formal sentence into informal sentence based on given instruction, and nothing else.
\textbf{</SYS>}
\end{tcolorbox}

\begin{tcolorbox}[title=General Informal Transfer Prompt, colback=gray!15, colframe=black!60]
Make the formal sentence into informal sentence.

\textbf{Informal Sentence} : {SEQUENCE}

\textbf{Formal Sentence} : 
\end{tcolorbox}

\begin{tcolorbox}[title=Review Domain Prompt, colback=gray!15, colframe=black!60]
Convert the following formal sentence into an informal review-style expression. Transform descriptive language into personal, evaluative expressions with clear positive/negative sentiment. Use review-typical phrases and authentic personal reactions.

\textbf{Formal sentence}: {SEQUENCE}

Convert to informal review style:
\end{tcolorbox}

\begin{tcolorbox}[title=Conversational and Social Commentary Domain Prompt, colback=gray!15, colframe=black!60]
Convert the following formal sentence into casual social media/chat style language. Use contractions, slang, emoticons, and reactive expressions that show immediate emotional response. Include satirical or meme-like elements where appropriate.

\textbf{Formal sentence}: {SEQUENCE}

Convert to social media/chat style:
\end{tcolorbox}

\begin{tcolorbox}[title=Professional Communication Domain Prompt, colback=gray!15, colframe=black!60]
Convert the following formal sentence into professional but approachable informal language. Maintain respect while reducing formality, use warm relationship-building language, and keep sentiment constructive and collaborative.

\textbf{Formal sentence}: {SEQUENCE}

Convert to professional informal style:
\end{tcolorbox}

\begin{tcolorbox}[title=Real-time \& Retrospective Expression, colback=gray!15, colframe=black!60]
Convert the following formal sentence into immediate, real-time reaction language. Use strong emotional expressions, short punchy sentences, reactive words, and present-tense immediacy.

\textbf{Formal sentence}: {SEQUENCE}

Convert to real-time reaction style:
\end{tcolorbox}

\section{Error Case Examples}
\label{appx:error}
\subsection{GPT-4.1-nano}

\newcolumntype{L}[1]{>{\raggedright\arraybackslash}m{#1}}

\begin{table*}[ht]
\centering
\small
\begin{tabular}{@{}clL{10.5cm}@{}}
\toprule
\textbf{\#} & \textbf{Category} & \textbf{Example} \\
\midrule
1 & Character Shift & consensual $\rightarrow$ consantal \\
\cmidrule{1-3}
2 & Entity Shift & good god how old are you $\rightarrow$ One may wonder about the age of the individual in question. \\
\cmidrule{1-3}
3 & Number Shift & 0.50 $\rightarrow$ by 0.52\% \\
\cmidrule{1-3}
4 & Addition (Entity-based) &
"I Have Nothing" by Jennifer Hudson $\rightarrow$ The song "I Have Nothing" by Jennifer Hudson is performed in the context of a musical expression. \vspace{1mm} \newline
For me it's definitely Jessica Alba and Angelina Jolie... $\rightarrow$ Jessica Alba and Angelina Jolie are the two actresses I consider to be the most appealing. \\
\cmidrule{1-3}
5 & Bias Injection &
In May, in his role as peace envoy, Blair met the education minister of the United Arab Emirates. $\rightarrow$ In May, shady Blair, just messing around as a peace envoy, met up with the UAE's education minister. \\
\cmidrule{1-3}
6 & Unresolved Aleatoric Uncertainty &
id have to say... Will Ferrell. $\rightarrow$ In my opinion, the actor Will Ferrell would be most representative. \vspace{1mm} \newline
or, what about blue and green? $\rightarrow$ Alternatively, what might be considered utilizing the colors blue and green? \\
\bottomrule
\end{tabular}
\caption{Examples of typical generation errors of GPT-4.1-nano categorized by type.}
\end{table*}
\clearpage

\subsection{T5-large}

\begin{table*}[ht]
\centering
\small
\begin{tabular}{@{}clL{10.5cm}@{}}
\toprule
\textbf{\#} & \textbf{Category} & \textbf{Example} \\
\midrule
1 & Deletion & have an equal opportunity to bid for capacity $\rightarrow$ bid for capacity \\
\cmidrule{1-3}
2 & Key Shift & will close in September 2011 $\rightarrow$ in the near future
\vspace{1mm} \newline
No way im 5`4 and he`s 6`2 $\rightarrow$ I am 5'4 and he is 6'2
\vspace{1mm} \newline
The sources I have listed below have all the information $\rightarrow$ i have listed all the sources\\
\cmidrule{1-3}
3 & Truncation & "Money that went to the armed forces that could have been or should have been spent on health and education, social services, was basically squandered. In any case the time is right now for democracy, for the people of of Guinea to get the elections they were hoping for," he added. $\rightarrow$ Money that went to the armed forces that could have been or should have been spent on health and education, social services, was basically squandered. In any case, the time.\\
\cmidrule{1-3}
4 & Copy \& Paste & It was trying so hard to be the next great American horror film. $\rightarrow$ It was trying so hard to be the next great American horror film. \\
\bottomrule
\end{tabular}
\caption{Examples of typical generation errors of T5 categorized by type.}
\end{table*}


\begin{thebibliography}{99}

\bibitem[Pavlick et~al.(2016)]{pavlick-tetreault-2016-empirical}
Pavlick, Ellie and Tetreault, Joel. 2016. An Empirical Analysis of Formality in Online Communication. Transactions of the Association for Computational Linguistics. \doi{10.1162/tacl_a_00083}.

\bibitem[Rao et~al.(2018)]{rao-tetreault-2018-dear}
Rao, Sudha and Tetreault, Joel. 2018. Dear Sir or Madam, May I Introduce the GYAFC Dataset: Corpus, Benchmarks and Metrics for Formality Style Transfer. Proceedings of the 2018 Conference of the North American Chapter of the Association for Computational Linguistics: Human Language Technologies, Volume 1 (Long Papers). \doi{10.18653/v1/N18-1012}.

\bibitem[Briakou et~al.(2021)]{briakou-etal-2021-ola}
Briakou, Eleftheria and Lu, Di and Zhang, Ke and Tetreault, Joel. 2021. Ol\'a, Bonjour, Salve! XFORMAL: A Benchmark for Multilingual Formality Style Transfer. Proceedings of the 2021 Conference of the North American Chapter of the Association for Computational Linguistics: Human Language Technologies. \doi{10.18653/v1/2021.naacl-main.256}.

\bibitem[Mukherjee et~al.(2024)]{mukherjee2024textstyletransferintroductory}
Sourabrata Mukherjee and Ondrej Duek. 2024. Text Style Transfer: An Introductory Overview. arXiv. \url{https://arxiv.org/abs/2407.14822}.

\bibitem[Liu et~al.(2024)]{liu-etal-2024-step}
Liu, Pusheng and Wu, Lianwei and Wang, Linyong and Guo, Sensen and Liu, Yang. 2024. Step-by-Step: Controlling Arbitrary Style in Text with Large Language Models. Proceedings of the 2024 Joint International Conference on Computational Linguistics, Language Resources and Evaluation (LREC-COLING 2024). \url{https://aclanthology.org/2024.lrec-main.1328/}.

\bibitem[Lai et~al.(2024)]{Lai2024StyleSpecificNF}
Wen Lai and Viktor Hangya and Alexander Fraser. 2024. Style-Specific Neurons for Steering LLMs in Text Style Transfer. ArXiv. \url{https://api.semanticscholar.org/CorpusID:273023003}.

\bibitem[Saakyan et~al.(2024)]{saakyan-muresan-2024-iclef}
Saakyan, Arkadiy and Muresan, Smaranda. 2024. ICLEF: In-Context Learning with Expert Feedback for Explainable Style Transfer. Proceedings of the 62nd Annual Meeting of the Association for Computational Linguistics (Volume 1: Long Papers). \doi{10.18653/v1/2024.acl-long.854}.

\bibitem[Toshevska et~al.(2025)]{toshevska-etal-2025-style}
Toshevska, Martina and Kalajdziski, Slobodan and Gievska, Sonja. 2025. Style Knowledge Graph: Augmenting Text Style Transfer with Knowledge Graphs. Proceedings of the Workshop on Generative AI and Knowledge Graphs (GenAIK). \url{https://aclanthology.org/2025.genaik-1.13/}.

\bibitem[Heylighen(1970)]{Heylighen_and_Dewaele}
Heylighen, Francis. 1970. Formality of Language: definition, measurement and behavioral determinants.

\bibitem[Biber(1988)]{Biber_1988}
Biber, Douglas. 1988. Variation across Speech and Writing. Cambridge University Press.

\bibitem[Biber et~al.(2009)]{Biber_Conrad_2009}
Biber, Douglas and Conrad, Susan. 2009. Register, Genre, and Style. Cambridge University Press.

\bibitem[Lai et~al.(2024)]{lai-etal-2024-style}
Lai, Wen and Hangya, Viktor and Fraser, Alexander. 2024. Style-Specific Neurons for Steering LLMs in Text Style Transfer. Proceedings of the 2024 Conference on Empirical Methods in Natural Language Processing. \doi{10.18653/v1/2024.emnlp-main.745}.

\bibitem[Mukherjee et~al.(2025)]{mukherjee-etal-2025-evaluating}
Mukherjee, Sourabrata and Ojha, Atul Kr. and McCrae, John Philip and Dusek, Ondrej. 2025. Evaluating Text Style Transfer Evaluation: Are There Any Reliable Metrics?. Proceedings of the 2025 Conference of the Nations of the Americas Chapter of the Association for Computational Linguistics: Human Language Technologies (Volume 4: Student Research Workshop). \doi{10.18653/v1/2025.naacl-srw.41}.

\bibitem[Yang et~al.(2025)]{yang2025steeringlargelanguagemodels}
Xinchen Yang and Marine Carpuat. 2025. Steering Large Language Models with Register Analysis for Arbitrary Style Transfer. arXiv. \url{https://arxiv.org/abs/2505.00679}.

\bibitem[Dementieva et~al.(2023)]{dementieva-etal-2023-detecting}
Dementieva, Daryna and Babakov, Nikolay and Panchenko, Alexander. 2023. Detecting Text Formality: A Study of Text Classification Approaches. Proceedings of the 14th International Conference on Recent Advances in Natural Language Processing. \url{https://aclanthology.org/2023.ranlp-1.31/}.

\bibitem[Ostheimer et~al.(2023)]{ostheimer2023standardizationvalidationtextstyle}
Phil Ostheimer and Mayank Nagda and Marius Kloft and Sophie Fellenz. 2023. A Call for Standardization and Validation of Text Style Transfer Evaluation. arXiv. \url{https://arxiv.org/abs/2306.00539}.

\bibitem[Kong et~al.(2025)]{kong-etal-2025-neuron}
Kong, Chaona and Liu, Jianyi and Tang, Yifan and Zhang, Ru. 2025. Neuron Activation Modulation for Text Style Transfer: Guiding Large Language Models. Findings of the Association for Computational Linguistics: ACL 2025. \doi{10.18653/v1/2025.findings-acl.403}.

\bibitem[Koh et~al.(2024)]{koh-etal-2024-llms}
Koh, Hyukhun and Kim, Dohyung and Lee, Minwoo and Jung, Kyomin. 2024. Can LLMs Recognize Toxicity? A Structured Investigation Framework and Toxicity Metric. Findings of the Association for Computational Linguistics: EMNLP 2024. \doi{10.18653/v1/2024.findings-emnlp.353}.

\bibitem[Pauli et~al.(2025)]{pauli-etal-2025-mind}
Pauli, Amalie Brogaard and Augenstein, Isabelle and Assent, Ira. 2025. Mind the Style Gap: Meta-Evaluation of Style and Attribute Transfer Metrics. Findings of the Association for Computational Linguistics: EMNLP 2025. \doi{10.18653/v1/2025.findings-emnlp.1175}.

\bibitem[La Quatra et~al.(2024)]{10.1145/3678179}
La Quatra, Moreno and Gallipoli, Giuseppe and Cagliero, Luca. 2024. Self-supervised Text Style Transfer Using Cycle-Consistent Adversarial Networks. ACM Trans. Intell. Syst. Technol.. \doi{10.1145/3678179}.

\bibitem[Hartmann et~al.(2023)]{hartmann-etal-2023-more}
Hartmann, Jochen and Heitmann, Mark and Siebert, Christian and Schamp, Christina. 2023. More than a Feeling: Accuracy and Application of Sentiment Analysis. International Journal of Research in Marketing.

\bibitem[Sun et~al.(2024)]{sun-etal-2024-toward}
Sun, Zhewei and Hu, Qian and Gupta, Rahul and Zemel, Richard and Xu, Yang. 2024. Toward Informal Language Processing: Knowledge of Slang in Large Language Models. Proceedings of the 2024 Conference of the North American Chapter of the Association for Computational Linguistics: Human Language Technologies (Volume 1: Long Papers). \doi{10.18653/v1/2024.naacl-long.94}.

\bibitem[Tyler et~al.(2005)]{Tyler01042005}
Joshua R. Tyler and Dennis M. Wilkinson and Bernardo A. Huberman. 2005. E-Mail as Spectroscopy: Automated Discovery of Community Structure within Organizations. The Information Society. \doi{10.1080/01972240590925348}.

\bibitem[Peterson et~al.(2011)]{petersonetal}
Peterson, Kelly and Hohensee, Matt and Xia, Fei. 2011. Email formality in the workplace: a case study on the Enron corpus.

\bibitem[Guerra et~al.(2014)]{10.1145/2556195.2556261}
Guerra, Pedro Calais and Meira, Wagner and Cardie, Claire. 2014. Sentiment analysis on evolving social streams: how self-report imbalances can help. Proceedings of the 7th ACM International Conference on Web Search and Data Mining. \doi{10.1145/2556195.2556261}.

\bibitem[Deng et~al.(2024)]{deng-etal-2024-text}
Deng, Zheye and Chan, Chunkit and Wang, Weiqi and Sun, Yuxi and Fan, Wei and Zheng, Tianshi and Yim, Yauwai and Song, Yangqiu. 2024. Text-Tuple-Table: Towards Information Integration in Text-to-Table Generation via Global Tuple Extraction. Proceedings of the 2024 Conference on Empirical Methods in Natural Language Processing. \doi{10.18653/v1/2024.emnlp-main.523}.

\bibitem[Hartmann et~al.(2023)]{HARTMANN202375}
Jochen Hartmann and Mark Heitmann and Christian Siebert and Christina Schamp. 2023. More than a Feeling: Accuracy and Application of Sentiment Analysis. International Journal of Research in Marketing. \doi{https://doi.org/10.1016/j.ijresmar.2022.05.005}.

\bibitem[Brown et~al.(1987)]{Brown_Levinson_1987}
Brown, Penelope and Levinson, Stephen C.. 1987. Politeness: Some Universals in Language Usage. Cambridge University Press.

\bibitem[Aithal et~al.(2021)]{aithal-tan-2021-positivity}
Aithal, Madhusudhan and Tan, Chenhao. 2021. On Positivity Bias in Negative Reviews. Proceedings of the 59th Annual Meeting of the Association for Computational Linguistics and the 11th International Joint Conference on Natural Language Processing (Volume 2: Short Papers). \doi{10.18653/v1/2021.acl-short.39}.

\bibitem[Culpeper(2011)]{Culpeper_2011}
Culpeper, Jonathan. 2011. Impoliteness: Using Language to Cause Offence. Cambridge University Press.

\bibitem[Brown et~al.(2020)]{brown2020languagemodelsfewshotlearners}
Tom B. Brown and Benjamin Mann and Nick Ryder and Melanie Subbiah and Jared Kaplan and Prafulla Dhariwal and Arvind Neelakantan and Pranav Shyam and Girish Sastry and Amanda Askell and Sandhini Agarwal and Ariel Herbert-Voss and Gretchen Krueger and Tom Henighan and Rewon Child and Aditya Ramesh and Daniel M. Ziegler and Jeffrey Wu and Clemens Winter and Christopher Hesse and Mark Chen and Eric Sigler and Mateusz Litwin and Scott Gray and Benjamin Chess and Jack Clark and Christopher Berner and Sam McCandlish and Alec Radford and Ilya Sutskever and Dario Amodei. 2020. Language Models are Few-Shot Learners. arXiv. \url{https://arxiv.org/abs/2005.14165}.

\bibitem[Bender et~al.(2021)]{10.1145/3442188.3445922}
Bender, Emily M. and Gebru, Timnit and McMillan-Major, Angelina and Shmitchell, Shmargaret. 2021. On the Dangers of Stochastic Parrots: Can Language Models Be Too Big?. Proceedings of the 2021 ACM Conference on Fairness, Accountability, and Transparency. \doi{10.1145/3442188.3445922}.

\bibitem[Shani et~al.(2025)]{shani2025tokensthoughtsllmshumans}
Chen Shani and Dan Jurafsky and Yann LeCun and Ravid Shwartz-Ziv. 2025. From Tokens to Thoughts: How LLMs and Humans Trade Compression for Meaning. arXiv. \url{https://arxiv.org/abs/2505.17117}.

\bibitem[Han et~al.(2022)]{han-etal-2022-balancing}
Han, Xudong and Baldwin, Timothy and Cohn, Trevor. 2022. Balancing out Bias: Achieving Fairness Through Balanced Training. Proceedings of the 2022 Conference on Empirical Methods in Natural Language Processing. \doi{10.18653/v1/2022.emnlp-main.779}.

\end{thebibliography}
\end{document}